\documentclass{article}
\pdfoutput=1

\usepackage{microtype}
\usepackage{graphicx}
\usepackage{caption}
\usepackage{subcaption}
\usepackage{booktabs} 
\usepackage{todonotes}
\usepackage{diagbox}
\usepackage{mathtools}
\usepackage{hyperref}
\hypersetup{colorlinks,allcolors=black}

\usepackage{amssymb}
\usepackage{amsmath}
\usepackage{xcolor}

\definecolor{gray}{HTML}{F5F5F5}
\definecolor{hl}{HTML}{E0E0E0}

\usepackage[accepted]{icml2020}

\begin{document}

\twocolumn[
\icmltitle {Untapped Potential of Data Augmentation: A Domain Generalization Viewpoint}  

\icmlsetsymbol{equal}{*}

\begin{icmlauthorlist}
\icmlauthor{Vihari Piratla}{iitb}
\icmlauthor{Shiv Shankar}{umass}
\end{icmlauthorlist}

\icmlaffiliation{iitb}{Indian Institute of Technology, Bombay}
\icmlaffiliation{umass}{University of Massachusetts, Amherst}

\icmlcorrespondingauthor{Vihari Piratla}{viharipiratla@gmail.com}

\icmlkeywords{Robustness, spurious correlations, domain generalization, Out-of-domain robustness, Data augmentation}

\vskip 0.3in
]

\printAffiliationsAndNotice{}  
\begin{abstract}
 Data augmentation is a popular pre-processing trick to improve generalization accuracy. It is believed that by processing augmented inputs in tandem with the original ones, the model learns a more robust set of features which are shared between the original and augmented counterparts. However, we show that is not the case even for the best augmentation technique. In this work, we take a Domain Generalization viewpoint of augmentation based methods. This new perspective allowed for probing overfitting and delineating avenues for improvement.
 Our exploration with the state-of-art augmentation method provides evidence that the learned representations are not as robust even towards distortions used during training. This suggests evidence for the untapped potential of augmented examples.
\end{abstract}

\section{Introduction}
Contemporary learning algorithms demonstrate strong performance, even surpassing humans at times, when training and testing on similar distributions. Notwithstanding performance under such setting, they are far from human level robustness when evaluated under data shifts~\citep{GeirhosTR18,HendrycksT19,MuG19MNIST-c}. This problem is of central focus in learning distributionally robust models. 
While the related problem of robustness to imperceptible adversarial examples has received much larger interest ~\citep{Madry18}; there has been an increasing push toward expanding the definition of robustness to include naturally occurring corruptions~\citep{Engstrom19a}. This is especially so because best defenses against, the narrow focused, adversarial examples does much worse with robustness to natural corruptions~\citep{HendrycksAugMix2019}.

There is a growing interest in building systems with better out-of-domain generalization performance also called Distributional Robustness (DR). DR is pursued under various fronts: (a) imposing inductive biases that penalize spurious correlations more ~\citep{WangGL19} (b) through employing augmentation or optimization techniques to weed out known data-overfitting features of the dataset ~\citep{SagawaPT20,Geirhos18Texture} (c) general dataset-agnostic data augmentation ~\citep{HendrycksAugMix2019,RusakSZ20}. Arguably, general augmentation methods are more scalable. Robustness through general augmentation is the problem of our interest in this work.


Due to different manifestations of the DR problem, the research in this direction is somewhat fragmented. The objective of robustness to domain shifts is the common theme in domain generalization~\citep{ShankarPC18,CarlucciDB19,PiratlaNS20}, distributional robustness ~\citep{HendrycksAugMix2019,RusakSZ20}, identifying and mitigating dataset biases~\citep{GururanganSL18}. DR is a general version of the Domain Generalization (DG) problem. DG functions under the setting where the train data is drawn from multiple sources along with annotation of source id for each example with the objective of better generalization to unseen domains. Difference between DG and DR are superficial, the following assumptions of former are relaxed in the latter (1) the train data does not necessarily be pooled from multiple sources (2) annotation of the domain label per example could be missing. In this work, we borrow lessons from the DG line of work to emphasize the untapped potential of augmentations directed at improving distribution robustness.


Data augmentation technique is widely adopted for image preprocessing and has recently been shown to improve out-of-domain robustness~\citep{HendrycksAugMix2019}. It is, however, unclear how the augmented examples interact with the clean examples. Training under data augmentation resembles multi-source training of DG. An ideal DG algorithm exploits the train time domain variation so as to learn a hypothesis that is better equipped at generalizing to new domains. The Expected Risk Minimization (ERM) baseline on the other hand does not attend to the domain boundaries and yields bad domain-shift robustness owing to overfitting on the seen domains. Standard training on clean and augmented inputs is akin to this ERM baseline which is known to have the overfitting issue. We want to draw attention to the under-explored utility of domain generalization methods for even better robustness. The prime focus of this work is to explore augmentation techniques in the context of out-of-domain robustness. Although some of our claims may also carry to generalization error, it is beyond the scope of this work. 

In our exposition, we use the terms {\it domain, augmentation} and {\it input distribution} interchangeably. Data augmentations are drawn from label consistent transformations of the clean examples, the introduced data-shift in the train data from augmentations is no different from what is usually referred to as the domain in the DG literature.


We make the following contributions. 
\begin{itemize}
    \item {\bf Untapped potential}: We show that the standard augmentation (including state-of-art) methods under-utilize augmented examples by overfitting on them. 
    \item {\bf Future direction}: We note that there is a broad scope for improving augmentations for even better robustness and conclude with a discussion of future line of work that target the observed domain-overfitting patterns. 
\end{itemize}

\section{Untapped Potential of Augmentations}
In this section, we provide evidence of augmentation overfit by systematically exploring a recent state-of-art augmentation method.
As a case study, we use models trained with \emph{AugMix}~\citep{HendrycksAugMix2019} when trained on CIFAR-10, CIFAR-100 and ImageNet across different network architectures. 

Augmentation is a standard trick employed to improve generalization, dominantly in image applications. In the extreme case of catastrophic overfitting of the augmentations, the augmented examples cannot help generalize better on the original examples. On the other extreme, in the ideal scenario, we expect the algorithm to draw what is common between the clean and augmented examples without having to employ any specific features for either clean or augmented data. Vanilla augmentation need not lead to the ideal scenario of learning common features between clean and augmented inputs. 
For example, \citet{VasiljevicCS16} report that train-time blur augmentations do not generalize to unseen blurs. Furthermore multiple DG studies ~\citep{Motiian17, Ghifary15} show that train data containing instances under multiple rotations does not generalize to unseen rotations. In practice, standard augmentation falls in between the two extremes of catastrophic overfitting and perfect parameter sharing. 


We pose the question of how much feature sharing occurs between the clean and augmented examples with AugMix using measures borrowed from the DG literature. In section~\ref{sec:disc}, we probe how domain invariant are the representations obtained from various layers. Section~\ref{sec:csd} employs a recent common-specific decomposition strategy proposed in ~\citet{PiratlaNS20} to identify any augmentation-specific (overfitting) components in the model weights. Finally in section~\ref{sec:generalization}, we make a more controlled evaluation of the generalization to augmentations of varying severity levels. 

\subsection{Domain Divergence Measure}
\label{sec:disc}
Domain overfit can be qualitatively measured by looking at how transferable the parameters are between the train domains. The seminal paper on domain adaptation: ~\citet{Ben-David06}, proved an upper bound on generalization gap between any two domains in terms of a domain divergence metric. Equation~\ref{eqn:dom_div} provides this metric for a given hypothesis class $\mathcal{H}$ and source and target distributions: $\mathcal{S}, \mathcal{T}$ with their respective populations: $n, n\prime$.


\begin{align}
  d_{\mathcal{H}}(\mathcal{S},\mathcal{T}) = 
  & 2(1 - min_{\eta\in \mathcal{H}}\{ \frac{1}{n}\Sigma_{i=1}^nI[\eta(x_i)=0] \nonumber \\
  & + \frac{1}{n\prime}\Sigma_{i=n+1}^{n+n\prime}I[\eta(x_i)=1]\}) \label{eqn:dom_div}
\end{align}


Intuitively, the domain divergence would be low when the hypothesis class induced by the learned representations do not allow for domain prediction i.e. the representations should be domain invariant. Since it is hard to compute the divergence measure exactly, a proxy measure, accuracy of a trained discriminator proposed in ~\citet{Ganin16}, is adopted. We train a domain discriminator to discriminate augmented examples from clean examples. Higher the accuracy of the domain discriminator, greater is the domain overfit.

We probe for domain invariance of the representations learned by AugMix on CIFAR and ImageNet datasets. We use representations from two different layers: the penultimate and antepenultimate layers, penultimate layer is the layer before the softmax layer. 
The train data for the discriminator is collated from the representations of the clean ($x_c$) and augmented ($x_a$) examples along with their domain assignment: $\bigcup_{i}\{x_{ci}, 0\}\cup\{x_{ai}, 1\}$. A linear discriminator is then trained on 40,000 examples with equal proportion of clean and augmented. 
If the model learns generalizable common features, then information related to the augmentation's distortion should be minimal. 
On the other hand if the model relies on domain specific feature, that information will be present in the representation layers of the model. The same information can be used to correctly identify the domain of the input sample.

Table~\ref{tab:disc} shows the discriminator's performance for a range of models trained with AugMix. We report discrimination accuracy on unseen test examples that are similarly collected as train and shown in parenthesis is the train accuracy.
The domain predictive accuracy in the penultimate layer is close to random, however, in just the neighboring antepenultimate layer it is possible to perfectly discern if the representation is from a clean or augmented example. The prevalence of domain identifying information up until this layer is indicative of shallow parameter sharing between augmentations and clean examples. This highlights the need for measures that promote higher parameter sharing between augmentations and original instances. 

\begin{table*}
    \centering
    \begin{tabular}[b]{|l|r|r|r|r|r|}
      \hline
      & \multicolumn{2}{|c|}{CIFAR-10} & \multicolumn{2}{|c|}{CIFAR-100} & \multicolumn{1}{|c|}{ImageNet} \\\hline
      \diagbox{Layer}{Arch} & AC & WRN & AC & WRN & ResNet-50 \\\hline
      PL    & 50.2 (52.8) & 51.9 (52.3) & 52.3 (52.1) & 51.0 (50.8) & 54.5 (57.4) \\
      APL   & 100 (100) & 85.5 (91.8) & 100 (100) & 84.8 (86.6) & 76.8 (84.0)\\
      \hline
    \end{tabular}
  \caption{Test and train domain discrimination accuracy (train accuracy shown in brackets) on CIFAR-10, CIFAR-100 and ImageNet. PL and APL stands for penultimate and antepenultimate layers. AC and WRN denote AllConv and WideResNet architecture respectively.}
  \label{tab:disc}
\end{table*}

\begin{table}
    \centering
    \begin{tabular}[b]{|l|r|r|r|}
      \hline
      \diagbox{Dataset}{Arch} & AC & WRN & RN-50\\\hline
      CIFAR-10    & 0.6  & 0.4 & - \\
      CIFAR-100   & 0.5  & 0.2 & - \\
      ImageNet    & -    & -   &  0.5 \\\hline
    \end{tabular}
  \caption{Ratio of norm of specific components to common components, smaller the better, for CIFAR-10, CIFAR-100 with AllConv (AC) and WideResNet (WRN) architecture, ImageNet with ResNet-50 (RN-50) architecture.}
  \label{tab:csd}
\end{table}

\subsection{Common vs Specialized Components of the Classifier}
\label{sec:csd}
When training on multi-domain data, we desire to retain only the components of the classifier that rely on features that are common between the domains. This concept is related but different from domain-invariant representations. While domain-invariant representations require that the features are invariant between the domains, the decomposition method of ~\citet{PiratlaNS20} encourages features of consistent label-correlation between the domains. The latter thereby is less restrictive than domain invariant features. The presence of domain-specific components in the classifier hurts out-of-domain generalization and when fixed can readily translate to even better robustness~\citep{PiratlaNS20}. Further, by employing the decomposition procedure from their work and comparing the support of domain specific component vs the common component of the classifier, we can qualitatively comment on the robustness strength. 

In order to study if AugMix suffers from the presence of any domain specific components, we employ the common-specific decomposition on the trained checkpoints. We obtain penultimate layer representations for a randomly sampled 20,000 train examples of original and augmented images each. We then obtain optimal content-label classifier individually for clean and augmented instances. These are denoted as $w_{clean}, w_{aug}$ respectively. We are interested in decomposing these parameters in to a linear combination of common ($w_c$) and domain-varying ($w_s$) component accompanied by domain-specific combination parameter ($\gamma_{clean}, \gamma_{aug}$). This requires solving the following constrained problem shown in Equation block~\ref{eqn:csd_problem} \footnote{See theorem 1 of ~\citet{PiratlaNS20}}.

\begin{align}
  w_{clean} = w_c + \gamma_{clean}w_s \nonumber \\
  w_{aug} = w_c + \gamma_{aug}w_s \nonumber \\ 
  w_c \perp w_s \label{eqn:csd_problem}
\end{align}

Note from the decomposition problem that (1) contribution of the common component $w_c$ to each of $w_{clean}, w_{aug}$ is the same, and (2) the contribution of specific component $w_s$ varies.
In the ideal case when the representation contains only features of consistent label correlations between domains, then the domain specific components ($\gamma_{clean}w_s, \gamma_{aug}w_s$) are diminutive compared to the common component ($w_c$). On the other hand when the representations contain features that favor only one of the two domains, it manifests in strong domain specific components.

In Table~\ref{tab:csd}, we report the ratio of norms of specific and common components over a range of models trained with AugMix, expression for the reported measure shown below:
$$
\frac{\|[\gamma_{clean}w_s, \gamma_{aug}w_s]\|}{\|[w_c, w_c]\|}    
$$
where $[\cdot]$ denotes concatenation of the vectors and $\|\cdot\|$ represents the Frobenius norm.

Ideally the ratio is expected to be very close to zero as the specific components are negligible. However, for a range of AugMix trained models, the ratio is significantly non-zero implying that there is non-negligible support for specific components compared to the common component. This strongly suggests the scope for better robustness when fixed\footnote{However in our case, post-processing linear classifier of the checkpoints to only retain the common component worsened the mean corruption error. Would be more interesting to evaluate the decomposition during the train time following ~\citet{PiratlaNS20}}. Interestingly, note that the corruption error (not shown here) and the specific-common ratio are inversely proportional on same dataset but different architecture. 

\subsection{Controlled Evaluation of Distributional Robustness}
\label{sec:generalization}
In this section, inspired from \citet{GeirhosTR18, VasiljevicCS16}, we make a controlled evaluation of the AugMix trained models in order to objectively measure domain sensitivity. AugMix allows for several knobs on the train time augmentations; Of our particular interest are (1) {\it mixing coefficient} that combines the augmented example with the original example (2) {\it severity level} of distortions for input transformation. We make a more modest evaluation on the test set using only the seen distortions but with differing distortion severity and with or without mixing with clean examples. 

Table~\ref{tab:gen} summarizes our findings. Without mixing means we evaluate on the augmented example directly. AugMix draws several samples from the convex combination of clean and distorted examples, and thereby we expect generalization to any convex combination of clean and augmented examples including either extremes. However, it is surprising that we found consistent drop in accuracy with the default severity level of 3 and when evaluated on an endpoint: distorted input. Also, we draw attention to the drop in accuracy when using severity level of 5 just outside of the train time value of 3 \footnote{Augmix severity scale is from 0 to 10}. These observations highlight the fragile robustness of AugMix.  

\begin{table}[htb]
\begin{tabular}{|l|r|r|r|r|}
\hline
      & \multicolumn{2}{|c|}{CIFAR-100} & \multicolumn{2}{|c|}{CIFAR-10}\\\hline
Test & \multicolumn{2}{|c|}{71.2} & \multicolumn{2}{|c|}{92.6}   \\\hline
& Mix & wo Mix & Mix & wo Mix \\\hline
s=3 & 69.9 (0.1) & 65.4 (0.3) & 91.8 (0.1) & 88.9 (0.1) \\
s=5 & 66.8 (0.3) & 61.4 (0.4) & 90.1 (0.2) & 87 (0.1) \\\hline
\end{tabular}
\caption{Classification accuracy of AugMix trained on CIFAR-100 and CIFAR-10 when evaluated on seen distortions of varying severity level (rows) and with ({\it Mix}) or without mixing ({\it wo Mix}) with clean example. The row corresponding to {\it Test} denotes performance on clean test set.}
\label{tab:gen}
\end{table}

\section{Related Work}

\paragraph{Domain Generalization}

Domain generalization refers to zero-shot adaptation to examples from unseen new domains. Building on \citet{Ben-David06} insight; a plethora of methods based on minimizing some form of domain divergence have been proposed \citep{Ganin16}. Other methods for domain generalization include parameter decomposition \citep{ECCV12_Khosla, PiratlaNS20} and meta-learning \citep{BalajiSC18,LiYSH18a}.

\paragraph{Data Augmentation} Various techniques like random erasure \citep{Zhong0KL020}, random replacement \citep{TakahashiMU18}, noise patching \citep{LopesYPGC19}, and image interpolation \citep{TokozumeUH18} have been explored for creation of augmented data samples. Both \citet{Madry18} and \citet{ShankarPC18} are versions of creating augmented examples using input gradient. \citet{XieHLL19} used augmentation in semi-supervised teacher student framework. 


\section{Discussion}
Standard training on clean and augmented examples combined need not realize the full potential of augmentations. Deep neural networks can learn unexpected properties and overfit on augmentations without delivering on the desired generalization. 

AugMix~\citep{HendrycksAugMix2019}, samples distortions from a large pool making it harder to overfit on a non-fixed set of distortions. 
The intent is to force learning of only features that transfer between clean and augmented examples. However, we present evidence in our work that contradicts this idealistic scenario of feature or parameter sharing with augmentations. On a range of datasets and architectures, AugMix employs specialized features for augmented examples as indicated by the domain divergence measure and common-specific decomposition of the classification layer. The utilization of specialized features could hinder out-of-domain generalization. 


Controlled evaluation on distortions obtained from slightly different distortion sampling parameters expose the fragile robustness to unseen but easy distortions. Furthermore the model retains sensitivity to training parameters. The mixing operation used in Augmix would lead one to expect that the model is robust on the simplex between clean data and its augmentations. However contrary to expectations even on the training augmentations, one sees significant difference between different mixing patterns. The fact that the models are not as robust as believed, suggests there is still significant scope of improvement from the way augmentations are currently utilized.

We envision future work targeting the patterns observed in this work. Mitigation of specific components in the representations can be achieved by adopting methods from~\citet{PiratlaNS20,SanyalPP2020}. Parameter sharing can be further promoted through a systematic study of domain invariant networks~\citep{Ganin16}.


\bibliography{main}

\begin{thebibliography}{28}
\providecommand{\natexlab}[1]{#1}
\providecommand{\url}[1]{\texttt{#1}}
\expandafter\ifx\csname urlstyle\endcsname\relax
  \providecommand{\doi}[1]{doi: #1}\else
  \providecommand{\doi}{doi: \begingroup \urlstyle{rm}\Url}\fi

\bibitem[Balaji et~al.(2018)Balaji, Sankaranarayanan, and
  Chellappa]{BalajiSC18}
Balaji, Y., Sankaranarayanan, S., and Chellappa, R.
\newblock Metareg: Towards domain generalization using meta-regularization.
\newblock In \emph{Advances in Neural Information Processing Systems 31: Annual
  Conference on Neural Information Processing Systems}, pp.\  1006--1016, 2018.

\bibitem[Ben-David et~al.(2006)Ben-David, Blitzer, Crammer, and
  Pereira]{Ben-David06}
Ben-David, S., Blitzer, J., Crammer, K., and Pereira, F.
\newblock Analysis of representations for domain adaptation.
\newblock In \emph{Proceedings of the 19th International Conference on Neural
  Information Processing Systems}, NIPS'06, 2006.
\newblock URL \url{http://dl.acm.org/citation.cfm?id=2976456.2976474}.

\bibitem[Carlucci et~al.(2019)Carlucci, D'Innocente, Bucci, Caputo, and
  Tommasi]{CarlucciDB19}
Carlucci, F.~M., D'Innocente, A., Bucci, S., Caputo, B., and Tommasi, T.
\newblock Domain generalization by solving jigsaw puzzles.
\newblock In \emph{Proceedings of the IEEE Conference on Computer Vision and
  Pattern Recognition}, pp.\  2229--2238, 2019.

\bibitem[Engstrom et~al.(2019)Engstrom, Gilmer, Goh, Hendrycks, Ilyas, Madry,
  Nakano, Nakkiran, Santurkar, Tran, Tsipras, and Wallace]{Engstrom19a}
Engstrom, L., Gilmer, J., Goh, G., Hendrycks, D., Ilyas, A., Madry, A., Nakano,
  R., Nakkiran, P., Santurkar, S., Tran, B., Tsipras, D., and Wallace, E.
\newblock A discussion of 'adversarial examples are not bugs, they are
  features'.
\newblock \emph{Distill}, 2019.
\newblock \doi{10.23915/distill.00019}.
\newblock https://distill.pub/2019/advex-bugs-discussion.

\bibitem[Ganin et~al.(2016)Ganin, Ustinova, Ajakan, Germain, Larochelle,
  Laviolette, Marchand, and Lempitsky]{Ganin16}
Ganin, Y., Ustinova, E., Ajakan, H., Germain, P., Larochelle, H., Laviolette,
  F., Marchand, M., and Lempitsky, V.
\newblock Domain-adversarial training of neural networks.
\newblock \emph{The Journal of Machine Learning Research}, 17\penalty0
  (1):\penalty0 2096--2030, 2016.

\bibitem[Geirhos et~al.(2018)Geirhos, Temme, Rauber, Sch{\"u}tt, Bethge, and
  Wichmann]{GeirhosTR18}
Geirhos, R., Temme, C.~R., Rauber, J., Sch{\"u}tt, H.~H., Bethge, M., and
  Wichmann, F.~A.
\newblock Generalisation in humans and deep neural networks.
\newblock In \emph{Advances in Neural Information Processing Systems}, pp.\
  7538--7550, 2018.

\bibitem[Geirhos et~al.(2019)Geirhos, Rubisch, Michaelis, Bethge, Wichmann, and
  Brendel]{Geirhos18Texture}
Geirhos, R., Rubisch, P., Michaelis, C., Bethge, M., Wichmann, F.~A., and
  Brendel, W.
\newblock Imagenet-trained {CNN}s are biased towards texture; increasing shape
  bias improves accuracy and robustness.
\newblock In \emph{International Conference on Learning Representations}, 2019.
\newblock URL \url{https://openreview.net/forum?id=Bygh9j09KX}.

\bibitem[Ghifary et~al.(2015)Ghifary, Bastiaan~Kleijn, Zhang, and
  Balduzzi]{Ghifary15}
Ghifary, M., Bastiaan~Kleijn, W., Zhang, M., and Balduzzi, D.
\newblock Domain generalization for object recognition with multi-task
  autoencoders.
\newblock In \emph{Proceedings of the IEEE international conference on computer
  vision}, pp.\  2551--2559, 2015.

\bibitem[Gururangan et~al.(2018)Gururangan, Swayamdipta, Levy, Schwartz,
  Bowman, and Smith]{GururanganSL18}
Gururangan, S., Swayamdipta, S., Levy, O., Schwartz, R., Bowman, S., and Smith,
  N.~A.
\newblock Annotation artifacts in natural language inference data.
\newblock In \emph{Proceedings of the 2018 Conference of the North {A}merican
  Chapter of the Association for Computational Linguistics: Human Language
  Technologies, Volume 2 (Short Papers)}, pp.\  107--112, New Orleans,
  Louisiana, June 2018. Association for Computational Linguistics.
\newblock \doi{10.18653/v1/N18-2017}.
\newblock URL \url{https://www.aclweb.org/anthology/N18-2017}.

\bibitem[Hendrycks \& Dietterich(2019)Hendrycks and Dietterich]{HendrycksT19}
Hendrycks, D. and Dietterich, T.
\newblock Benchmarking neural network robustness to common corruptions and
  perturbations.
\newblock \emph{Proceedings of the International Conference on Learning
  Representations}, 2019.

\bibitem[Hendrycks et~al.(2019)Hendrycks, Mu, Cubuk, Zoph, Gilmer, and
  Lakshminarayanan]{HendrycksAugMix2019}
Hendrycks, D., Mu, N., Cubuk, E.~D., Zoph, B., Gilmer, J., and
  Lakshminarayanan, B.
\newblock Augmix: A simple data processing method to improve robustness and
  uncertainty.
\newblock \emph{arXiv preprint arXiv:1912.02781}, 2019.

\bibitem[Khosla et~al.(2012)Khosla, Zhou, Malisiewicz, Efros, and
  Torralba]{ECCV12_Khosla}
Khosla, A., Zhou, T., Malisiewicz, T., Efros, A., and Torralba, A.
\newblock Undoing the damage of dataset bias.
\newblock In \emph{ECCV}, pp.\  158--171, 2012.

\bibitem[Li et~al.(2018)Li, Yang, Song, and Hospedales]{LiYSH18a}
Li, D., Yang, Y., Song, Y., and Hospedales, T.~M.
\newblock Learning to generalize: Meta-learning for domain generalization.
\newblock In \emph{Proceedings of the Thirty-Second {AAAI} Conference on
  Artificial Intelligence, (AAAI-18), the 30th innovative Applications of
  Artificial Intelligence (IAAI-18), and the 8th {AAAI} Symposium on
  Educational Advances in Artificial Intelligence (EAAI-18)}, pp.\  3490--3497,
  2018.

\bibitem[Lopes et~al.(2019)Lopes, Yin, Poole, Gilmer, and Cubuk]{LopesYPGC19}
Lopes, R.~G., Yin, D., Poole, B., Gilmer, J., and Cubuk, E.~D.
\newblock Improving robustness without sacrificing accuracy with patch gaussian
  augmentation.
\newblock \emph{CoRR}, abs/1906.02611, 2019.
\newblock URL \url{http://arxiv.org/abs/1906.02611}.

\bibitem[Madry et~al.(2018)Madry, Makelov, Schmidt, Tsipras, and
  Vladu]{Madry18}
Madry, A., Makelov, A., Schmidt, L., Tsipras, D., and Vladu, A.
\newblock Towards deep learning models resistant to adversarial attacks.
\newblock In \emph{International Conference on Learning Representations}, 2018.
\newblock URL \url{https://openreview.net/forum?id=rJzIBfZAb}.

\bibitem[Motiian et~al.(2017)Motiian, Piccirilli, Adjeroh, and
  Doretto]{Motiian17}
Motiian, S., Piccirilli, M., Adjeroh, D.~A., and Doretto, G.
\newblock Unified deep supervised domain adaptation and generalization.
\newblock In \emph{Proceedings of the IEEE International Conference on Computer
  Vision}, pp.\  5715--5725, 2017.

\bibitem[Mu \& Gilmer(2019)Mu and Gilmer]{MuG19MNIST-c}
Mu, N. and Gilmer, J.
\newblock Mnist-c: A robustness benchmark for computer vision.
\newblock \emph{arXiv preprint arXiv:1906.02337}, 2019.

\bibitem[Piratla et~al.(2020)Piratla, Netrapalli, and Sarawagi]{PiratlaNS20}
Piratla, V., Netrapalli, P., and Sarawagi, S.
\newblock Efficient domain generalization via common-specific low-rank
  decomposition.
\newblock \emph{arXiv preprint arXiv:2003.12815}, 2020.

\bibitem[Rusak et~al.(2020)Rusak, Schott, Zimmermann, Bitterwolf, Bringmann,
  Bethge, and Brendel]{RusakSZ20}
Rusak, E., Schott, L., Zimmermann, R., Bitterwolf, J., Bringmann, O., Bethge,
  M., and Brendel, W.
\newblock Increasing the robustness of dnns against image corruptions by
  playing the game of noise.
\newblock \emph{arXiv preprint arXiv:2001.06057}, 2020.

\bibitem[Sagawa* et~al.(2020)Sagawa*, Koh*, Hashimoto, and Liang]{SagawaPT20}
Sagawa*, S., Koh*, P.~W., Hashimoto, T.~B., and Liang, P.
\newblock Distributionally robust neural networks.
\newblock In \emph{International Conference on Learning Representations}, 2020.
\newblock URL \url{https://openreview.net/forum?id=ryxGuJrFvS}.

\bibitem[Sanyal et~al.(2020)Sanyal, Torr, and Dokania]{SanyalPP2020}
Sanyal, A., Torr, P.~H., and Dokania, P.~K.
\newblock Stable rank normalization for improved generalization in neural
  networks and gans.
\newblock In \emph{International Conference on Learning Representations}, 2020.
\newblock URL \url{https://openreview.net/forum?id=H1enKkrFDB}.

\bibitem[Shankar et~al.(2018)Shankar, Piratla, Chakrabarti, Chaudhuri, Jyothi,
  and Sarawagi]{ShankarPC18}
Shankar, S., Piratla, V., Chakrabarti, S., Chaudhuri, S., Jyothi, P., and
  Sarawagi, S.
\newblock Generalizing across domains via cross-gradient training.
\newblock \emph{arXiv preprint arXiv:1804.10745}, 2018.

\bibitem[Takahashi et~al.(2018)Takahashi, Matsubara, and Uehara]{TakahashiMU18}
Takahashi, R., Matsubara, T., and Uehara, K.
\newblock Data augmentation using random image cropping and patching for deep
  cnns.
\newblock \emph{CoRR}, abs/1811.09030, 2018.
\newblock URL \url{http://arxiv.org/abs/1811.09030}.

\bibitem[Tokozume et~al.(2018)Tokozume, Ushiku, and Harada]{TokozumeUH18}
Tokozume, Y., Ushiku, Y., and Harada, T.
\newblock Between-class learning for image classification.
\newblock In \emph{2018 {IEEE} Conference on Computer Vision and Pattern
  Recognition}, pp.\  5486--5494. {IEEE} Computer Society, 2018.

\bibitem[Vasiljevic et~al.(2016)Vasiljevic, Chakrabarti, and
  Shakhnarovich]{VasiljevicCS16}
Vasiljevic, I., Chakrabarti, A., and Shakhnarovich, G.
\newblock Examining the impact of blur on recognition by convolutional
  networks.
\newblock \emph{arXiv preprint arXiv:1611.05760}, 2016.

\bibitem[Wang et~al.(2019)Wang, Ge, Lipton, and Xing]{WangGL19}
Wang, H., Ge, S., Lipton, Z., and Xing, E.~P.
\newblock Learning robust global representations by penalizing local predictive
  power.
\newblock In \emph{Advances in Neural Information Processing Systems}, pp.\
  10506--10518, 2019.

\bibitem[Xie et~al.(2019)Xie, Hovy, Luong, and Le]{XieHLL19}
Xie, Q., Hovy, E.~H., Luong, M., and Le, Q.~V.
\newblock Self-training with noisy student improves imagenet classification.
\newblock \emph{CoRR}, abs/1911.04252, 2019.

\bibitem[Zhong et~al.(2020)Zhong, Zheng, Kang, Li, and Yang]{Zhong0KL020}
Zhong, Z., Zheng, L., Kang, G., Li, S., and Yang, Y.
\newblock Random erasing data augmentation.
\newblock In \emph{The Thirty-Fourth {AAAI} Conference on Artificial
  Intelligence, {AAAI} 2020, The Thirty-Second Innovative Applications of
  Artificial Intelligence Conference, {IAAI} 2020, The Tenth {AAAI} Symposium
  on Educational Advances in Artificial Intelligence, {EAAI} 2020}, pp.\
  13001--13008. {AAAI} Press, 2020.

\end{thebibliography}
\bibliographystyle{icml2020}
\end{document}